\pdfoutput=1

\documentclass[11pt]{article}

\usepackage[]{EMNLP2023}
\usepackage{url}
\usepackage{times}
\usepackage{latexsym}
\usepackage{caption}
\usepackage{subcaption}
\usepackage{graphicx} 
\usepackage[T1]{fontenc}

\usepackage[utf8]{inputenc}

\usepackage{microtype}

\usepackage{inconsolata}
\usepackage{booktabs}
\usepackage{tabularx}
\usepackage{adjustbox}
\usepackage{multirow}
\usepackage{amssymb}
\usepackage{xcolor}
\definecolor{darkgreen}{RGB}{0,100,0}
\usepackage{longtable}
\usepackage{colortbl}
\usepackage{amsmath}

\title{Know Your Audience: Do LLMs Adapt to Different Age and Education Levels?}

\author{Donya Rooein, Amanda Cercas Curry, Dirk Hovy  \\
    Bocconi University, Milan, Italy \\
  \texttt{donya.rooein@unibocconi.it}}

\begin{document}
\maketitle
\begin{abstract}
Large language models (LLMs) offer a range of new possibilities, including adapting the text to different audiences and their reading needs. But how well do they adapt?
We evaluate the readability of answers generated by four state-of-the-art LLMs (commercial and open-source) to science questions when prompted to target different age groups and education levels. To assess the adaptability of LLMs to diverse audiences, we compare the readability scores of the generated responses against the recommended comprehension level of each age and education group. 
We find large variations in the readability of the answers by different LLMs. 
Our results suggest LLM answers need to be better adapted to the intended audience demographics to be more comprehensible. They underline the importance of enhancing the adaptability of LLMs in education settings to cater to diverse age and education levels. 
Overall, current LLMs have set readability ranges and do not adapt well to different audiences, even when prompted. That limits their potential for educational purposes.

\end{abstract}

\section{Introduction}
\label{sec:intro}

\begin{figure}[ht!]
    \centering
    \includegraphics[width=\columnwidth]{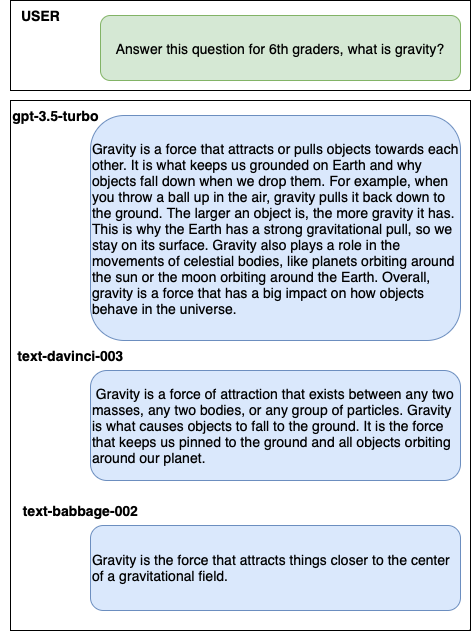}
    \caption{Example of prompting different LLMs for specific education levels. The gpt-3.5-turbo provides a longer response with examples.}
    \captionsetup{justification=centering}
    \label{fig:QA-models}
\end{figure}

``Where does the sun go at night?'' is a simple enough science question. However, the ``best'' answer depends on whether we respond to a toddler, a 6-year-old, or a high-schooler.
Large language models (LLMs) have been a significant breakthrough in natural language processing, enabling the modeling of complex linguistic phenomena and capturing factual and commonsense knowledge. However, despite their impressive performance, recent studies have shown that in many cases, they still need to learn how to provide appropriate answers for a given audience~\citet{qureshi2023chatgpt}. 

In this paper, we ask, ``How can an LLM adapt to an audience with a particular age and education level?''
Specifically, we test how well four of the most commonly used models (two commercial and two open-source) adapt to various age groups and education levels. We use widely-available readability metrics to assess whether a model's responses are well-adapted to the requested age and educational level.

LLMs like GPT-3~\citep{brown2020language} have become an essential tools in developing advanced dialogue systems that can effectively communicate with users. By leveraging the pre-trained parameters in LLMs, dialogue systems can generate a wide range of appropriate responses for a given context, enhancing the overall user experience.
These advanced dialogue systems can respond to a user's request in a natural and human-like manner, making them more effective at providing useful responses. And while LLMs are designed to adapt to different writing styles (casual vs.\ formal) and domains (email, blog, essay), their ability to adapt to different \textit{education levels} is much more limited. Moreover, The utilization of natural language to direct the outputs of LLMs through the technique of "prompting" has gained recognition as a practical design approach that has the potential to be accessible to individuals without expertise in AI~\cite{zamfirescu2023johnny}.

A large amount of literature in pedagogy points to the importance of age-appropriate reading~\citep{nguyen2021construction}. LLMs are used increasingly in the classroom, so their lack of adaptability can severely impact them. While LLMs can enhance the accessibility to education and learning materials for learners. Figure~\ref{fig:QA-models} shows different LLM responses to a science question for a given education level. The outputs of gpt-3.5-turbo and Text-Davinci-003 adapt to the question, whereas the others fail to adapt to the requested level and provide incorrect answers to the question. This example suggests that some LLMs can answer science questions with some caveats. 


Here, we prompt different models for answers considering age, education level, and learning styles (e.g., easy or difficult explanation). We use standard metrics to evaluate the readability and age-appropriateness of the LLM responses. We find that LLMs cannot follow even explicit prompts for age, education level, or difficulty level and produce text that does not match the readability levels recommended for that audience.

\paragraph{Contributions}
We make the following contributions to this paper:
\begin{enumerate}
    \item We show that current LLMs do not effectively adapt output to specific targets like age groups, education levels, and level of difficulty except for very advanced categories, making them ill-suited for educational purposes. 

    \item We also show, however, that common readability metrics are not completely reliable for determining the education level of LLM-generated responses.

    \item We release our data set of questions, responses, and metrics for further study.
    
\end{enumerate}

\section{Methodology}
\label{sec:appro}

Recent research~\citep{wang2019demographic} has shown that the demographics of social media users, including gender, age, and education level, can be identified with high accuracy from the linguistic patterns in their Twitter profiles and posts. Education level influences vocabulary, syntax, and other linguistic features that are not always easy to model with a single LLM. For example, an LLM trained in academic texts may need help generating language at a lower education level, where more straightforward vocabulary and shorter sentences are common. Similarly, an LLM trained on social media posts may need help to produce accurate and grammatically correct text at a higher education level, where more complex sentence structures and technical vocabulary are required. 

Our approach is as follows: 
(1) we collect a data set of 33,600 prompt-response pairs from five models over 4 runs, 
(2) we assess whether the readability of the responses adapts to what the audience we requested in the prompt, using reference readability metrics, and 
(3) we determine the extent of change in readability metrics by analyzing the range of variations observed.

\subsection{Readability Metrics}
\label{sssec:readability_metrics}
Among common metrics to measure text difficulty, we use the Flesch-Kincaid Reading Ease Index (FKRE) due to its suitability for shorter sentences and its reliance on surface features such as word and sentence length in a given text~\citet{choudhery2020readability}. 
It assigns a score between 1 and 100, where higher scores indicate greater readability. 
The formula for measuring FKRE formula is given by:

\begin{equation*}
\label{eq:fkre}
\begin{aligned}
\text{FKRE} = &\ 206.835 - 1.015 \times \left(\frac{\text{\#words}}{\text{\#sent.}}\right) \\
&\ - 84.6 \times \left(\frac{\text{\#syll.}}{\text{\#words}}\right)
\end{aligned}
\end{equation*}


Based on the FKRE Index, we can also derive the intended grade level (Flesch-Kincaid Grade Level, FKGL), and the level of difficulty of the text for reading (based on the previous two). 
We use these metrics to define our target groups because they are the most commonly used in the literature. They also offer a high granularity for the appropriateness of a text by age, grade level, and difficulty. These aspects allow us to assess the LLMs' readability along more than one dimension. 
Other commonly used metrics, like Gunning-Fog Index, Coleman-Liau Index, and Simple Measure of Gobbledygook (SMOG) Index~\citet{choudhery2020readability}, are more specialized and need long-form documents to compute them, which is not applicable in our case. 
Among these metrics, FKRE is the most basic. FKGL is based on a similar formula as FKRE, but instead of a score maps it into the grade level required to comprehend a given passage.\footnote{\url{https://readable.com/readability/flesch-reading-ease-flesch-kincaid-grade-level/}} 
Note that this is the grade system for the US education system.

While both Flesh-Kincaid metrics use the same units, i.e., word totals, syllable toals, and sentence totals, they apply different weightings to these units to map them into their outcome scores. Consequently, the two scores serve as indicators of specific age groups and education grades.

Finally, the difficulty level is a mapping from FKRE score ranges into seven categorical values (very easy, fairly easy, easy, medium, difficult, fairly difficult, very difficult; see Table \ref{tab:param_summary}). In our analysis, we further collapse these into three categories, easy (71--100), medium (60--70), and hard (0--59), by removing the modifiers. This grouping also produces clearer trends than the 7-point scale.
Table~\ref{tab:param_summary} the mapping of the age, education level, suggested readability score, and FKRE range. 

In our question prompts for the LLMs, we use the values of the first three metrics. The correspondence to FKRE scores allows us to check the produced text and see whether it respects those mappings.
In addition, we prompt LLMs also for categories that are common in the field of education
``kids'' (all groups under 18), ``adult'' (over 18), and for ``professional'' readers (FKRE 0--10).
A list of resulting example prompts is available in Table~\ref{tab:question-targets}.

\begin{table}[!t]
\centering
\begin{tabularx}{\columnwidth}{>{\hsize=0.7\hsize}X>{\hsize=1.3\hsize}X>{\hsize=0.9\hsize}X>{\hsize=1.1\hsize}X}
\hline
Age & FKGL &  Difficulty & FKRE range\\ 
\hline
11 & 5th grade & very easy & 90 - 100\\
11-12 & 6th grade & fairly easy & 80 - 90\\
12-13 & 7th grade & easy & 70 - 80\\
13-15 & 8th--9th grade & medium & 60 - 70\\
15-18 & 10th--12th grade & difficult & 50 - 60\\
18-19 & College & fairly difficult & 30 - 50\\
22-23 & College graduate & very difficult & 0 - 30\\
\hline
\end{tabularx}
\caption{Mapping of age and grading levels to difficulty and metrics\footnote{\url{https://clickhelp.com/software-documentation-tool/user-manual/flesch-reading-ease.html}}.}
\label{tab:param_summary}
\end{table}

\subsection{Data}
We specifically selected the science domain because it encompasses different levels of abstraction for a given concept and is part of all educational levels -- with varying levels of detail. 

We repeatedly generated $100$ science questions by prompting ChatGPT with \textit{``Generate 100 questions about science''}.
We carefully reviewed the full set of generated questions for uniqueness and diversity. We removed any duplicates and then selected 100 questions to balance out scientific fields and areas (e.g., math, physics, biology, chemistry, etc.).
The full list of the 100 final questions is in Appendix~\ref{appendix:questions}.

For each age and education category, we manually design templates for the prompting of the LLMs. Let {${\tau_\alpha, \tau_\beta, ...,\tau_k}$} be the set of target groups that we want to include in our questions. 

These variables encompass various values representing age and education level, and difficulty appropriateness. By incorporating these target groups, we encourage LLMs to generate responses tailored to the specified criteria. 

In our collected dataset, given a question $q_i$ and the desired target group $\tau_\alpha$, the corresponding adapted answer is $a_i$. To mitigate the variability in model outputs, we prompt the models at different times for each question and collect multiple answers. A sample record in our data set is as follows: \\

$(q_i,\tau_\alpha) \rightarrow (a_{i,\tau,1}, a_{i,\tau,2}, ..., a_{i,\tau,m})$


\begin{table}[!ht]
\centering
\begin{tabular}{p{0.7\columnwidth}c}
\toprule
\textbf{Prompt} & \textbf{Category}\\ 
\midrule
Answer this question, how do we measure the strength of earthquakes ? & No target \\ \hline
Answer this question for an \textcolor{red}{11-year-old}, how do we measure the strength of earthquakes? & \textcolor{red}{Age} \\ \hline
Answer this question for \textcolor{blue}{college graduates}, how do we measure the strength of earthquakes? & \textcolor{blue}{Education} \\ \hline
Answer this question in an  \textcolor{magenta}{easy-level explanation}, how do we measure the strength of earthquakes? &  \textcolor{magenta}{Difficulty} \\ \hline
Answer this question for an  \textcolor{darkgreen}{adult}, how do we measure the strength of earthquakes? &  \textcolor{darkgreen}{Group} \\ \bottomrule
\end{tabular}

\caption{A list of generated prompts for collecting responses from LLMs.}
\label{tab:question-targets}
\end{table}



\subsection{Systems Compared} 

We investigate the readability of LLM-generated answers by tailoring them to suit specific age groups, education levels, and difficulty levels. While there are limited LLMs available for fine-tuning to address science-related queries and generate corresponding responses, our primary objective was to ensure that the generated answers adapted to the requested parameters. 

To get a realistic assessment for use in education, we choose four common LLMs, two commercial ones, and two open-source models.
We compare the adaptability of OpenAI's ChatGPT\footnote{\url{https://openai.com/blog/chatgpt}}, \texttt{GPT-3 Da-Vinci-0003}~\cite{brown2020language}, 
\texttt{bigscience-T0}~\cite{sanh2021multitask} and Flan-T5~\cite{chung2022scaling}. We know that the outputs from models trained to follow user instructions generally produce better outputs than causal models. In this study, our main objective is to evaluate the readability of the generated responses. We assess the appropriateness and ease of comprehension of the model-generated responses for the target groups. This analysis allows us to gain insights into the readability level of the generated content and its suitability for different age groups and educational backgrounds.


\begin{figure}[ht!]
  \centering
  \begin{subfigure}{\columnwidth}
    \centering
    \includegraphics[width=\columnwidth]{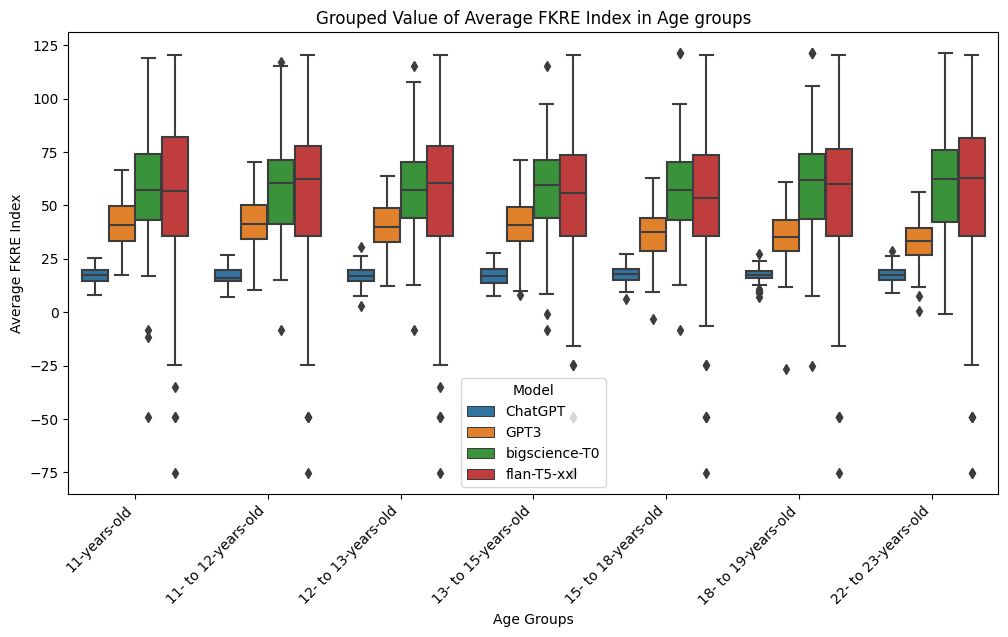}
    \caption{Age Groups}
    \label{fig:age}
  \end{subfigure}
  \hfill
  \begin{subfigure}{\columnwidth}
    \centering
    \includegraphics[width=\columnwidth]{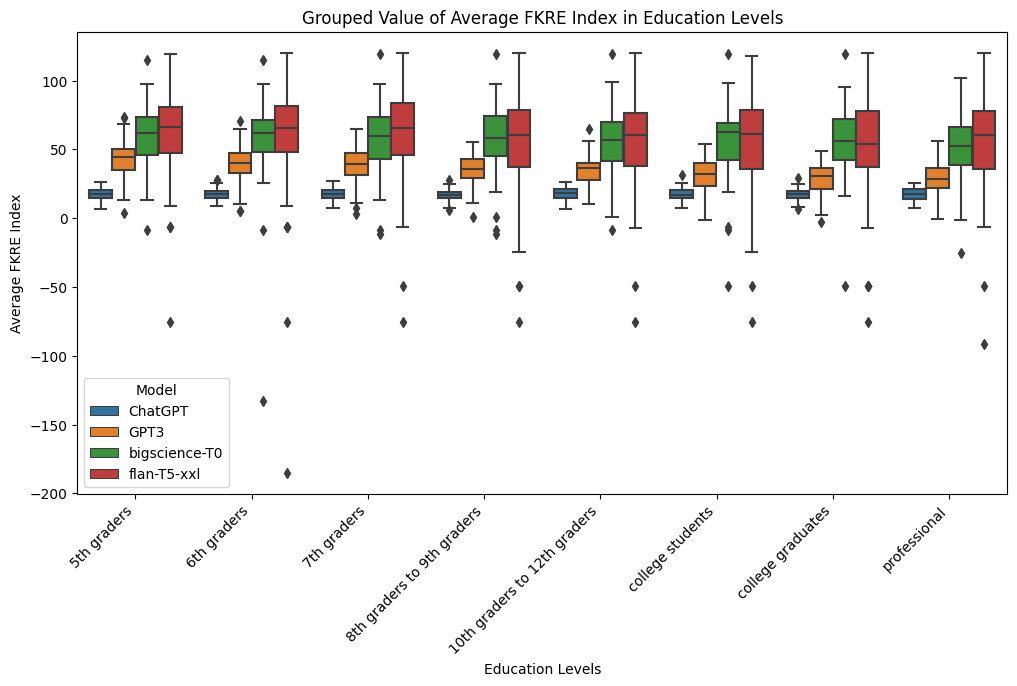}
    \caption{Education Levels}
    \label{fig:edu}
  \end{subfigure}
  \vskip\baselineskip
  \begin{subfigure}{\columnwidth}
    \centering
    \includegraphics[width=\columnwidth]{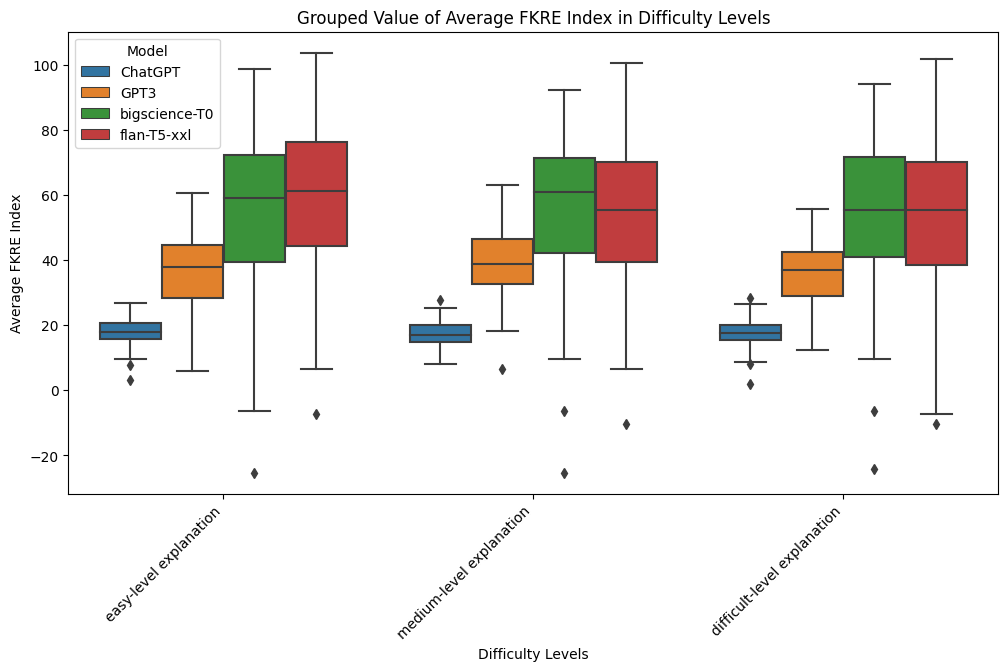}
    \caption{Difficulty levels}
    \label{fig:diff}
  \end{subfigure}
  \hfill
  \begin{subfigure}{\columnwidth}
    \centering
    \includegraphics[width=\columnwidth]{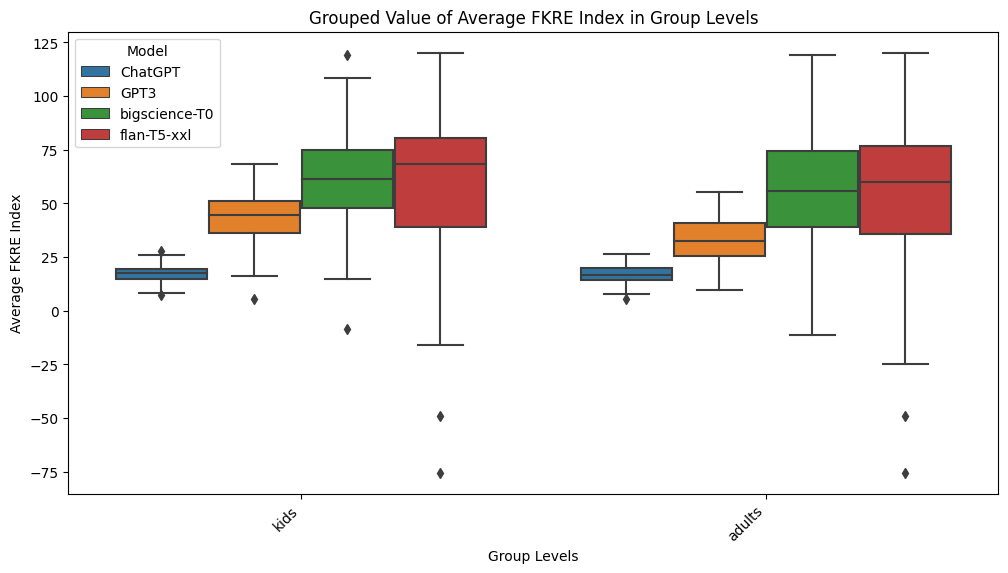}
    \caption{Group}
    \label{fig:group}
  \end{subfigure}
  \caption{Comparison of average FKRE ranges for different models in each target category}
  \label{fig:four_cat_FKRE}
\end{figure}

\section{Results} 
 
We first analyze the descriptive statistics of the generated answers across all target groups, using metrics such as average word count, sentence count, and Type-Token Ratio (TTR). Table~\ref{tab:dataset_description} reports detailed statistics on all the models' answers. 

Our analysis reveals that instruction-based models like ChatGPT and GPT3 tend to produce longer answers (more sentences and words) in response to the given questions than the other two models. 
We observe that the TTR, which measures vocabulary diversity, increases for each of the different models, with a peak for Bigscience-T0.
However, it does not vary significantly among targets within the same groups. For example, Table~\ref{tab:dataset_description} shows that for ChatGPT, the average TTR for an 11-year-old target is 0.637. This value decreases only slightly to 0.635 for the 22--23-year-old target. This observation suggests that vocabulary usage remains relatively consistent across these target groups within the same category.

\subsection{Readability Scores}
Ultimately, we want to know how well LLMs can adapt to different age groups and education levels. To assess this, we use the readability metrics described in Section \ref{sssec:readability_metrics} on the outputs of each set of LLM answers.

\begin{table*}[htb]
\center
\begin{tabular}{l|r|r|r|r}
\hline
Model & Avg. FKRE &Avg.\# sents & Avg.\# words & TTR \\ \hline
ChatGPT & $17.35$ &$4.9$ &  $103.4$ & $0.63$\\
GPT3 & $36.46$ &$6.4$ & $129.1$ & $0.70$\\
flan-T5-xxl & $56.74$ & $1.0$ & $11.6$& $0.83 $\\
Bigscience-T0 & $57.45$ &$1.0$ & $13.9$ & $0.93$\\
 \hline
\end{tabular}
\caption{Average summary statistics of evaluation responses. 
}
\label{tab:dataset_description}
\end{table*}

ChatGPT had an average FKRE index of 17.35, indicating that the average generated text tends to be rather difficult to read. It suggests that the generated text is highly challenging and is most suitable for individuals with a college graduate education level, or professionals.
GPT3 is more than twice as easy to read, with an average FKRE of 36.46. The answers also contain more sentences and more words than for chatGPT.
Flan-T5-xxl and Bigscience-T0 are even easier to read, though they tend to produce short answers with fewer words.

\begin{table*}[ht]
\centering
\begin{tabular}{lccccccc|c}
\toprule
\textbf{Age Target (years)} & 11 & 11--12 & 12--13 & 13--15 & 15--18 & 18--19 & 22--23 & \textbf{None}\\
\midrule
ChatGPT & 17.31 & 16.84 & 17.32 & 17.03 & 17.63 & 17.33 & \textbf{17.48} &17.25\\
GPT3 & 41.86 & 41.17 & 40.47 & 40.81 & 36.31 & \textbf{34.51} & 32.82 & 30.60\\
flan-T5-xxl &  56.47 & 55.73 & 55.64 & 51.58 & 49.90 & 53.67 & 57.93 &57.63\\
Bigscience-T0 & 57.63 & 58.75 & 57.78 & 58.71 & \textbf{57.44} & 60.04 & 61.25 &57.66\\
\midrule
\end{tabular}
\caption{Average FKRE index values for the intended/prompted audience, categorized by \textit{Age} and model. Performance within target range in bold.}
\label{tab:target-age}
\end{table*}
\begin{table*}[ht]
\centering
\resizebox{\textwidth}{!}{
\begin{tabular}{l|cccccp{1.3cm}p{1cm}c|c}
\toprule
\textbf{Target} & 5th grd. & 6th & 7th & 8th--9th  & 10th--12th & college students & college grads & professional & \textbf{None}\\
\midrule
ChatGPT & 17.62 & 17.54 & 17.68 & 16.92 & 17.90 & 17.35 & 17.19 & \textbf{17.46} & 17.25\\
GPT3 & 42.83 & 39.69 & 38.19 & 35.15 & 34.57 & 31.83 & 28.77 & 28.51 & 30.60 \\
flan-T5-xxl &  63.86 & 61.69 & 63.37 & 57.01 & \textbf{56.57} & 55.57 & 54.98 & 57.25 & 57.63\\
Bigscience-T0 & 59.02 & 57.87 & 57.59 & 58.59 & \textbf{56.38} & 56.44 & 57.42 & 51.22 & 57.66\\
\midrule
\end{tabular}
} 
\caption{Average FKRE index values for the intended/prompted audience, by \textit{Education} and model. Performance within target range in bold.}
\label{tab:target-edu}
\end{table*}

\begin{table}[ht]
\centering
\begin{tabular}{lcc|c}
\toprule
\textbf{Target} & Kid & Adult &\textbf{None}\\
\midrule
ChatGPT & 17.01 & 17.01 & 17.25\\
GPT3 & 43.09 & 32.53 & 30.60\\
flan-T5-xxl & 59.75 & 56.09 & 57.63\\
Bigscience-T0 & \textbf{60.24} & 55.99 & 57.66\\
\midrule
\end{tabular}
\caption{Average FKRE index values for the prompted audience, by \textit{Group} and model. Performance within target range in bold.}
\label{tab:target-group}
\end{table}

Figure~\ref{fig:four_cat_FKRE} shows box-and-whisker plots for the four models on the different prompt types.
We run a t-test to confirm that there is a significant difference between the calculated FKRE index among responses generated by the models with and without specifying a target. 
The None target group refers to the prompts that do not specify any particular target audience, while the other target groups correspond to prompts tailored to specific age, education, or difficulty levels. We find that most of the target group combinations do not show a significant difference in the means of the FKRE index. The p-values for these combinations are generally higher than the conventional threshold of 0.05, suggesting that the observed differences in the means are likely due to random chance. The detailed results of t-test are available in appendix~\ref{appendix:ttest}

ChatGPT generates responses that are in a narrow range of FKRE scores. 
The other models have increasingly larger ranges that do not vary much between prompt groups, however.
Overall, it seems each model has a particular ``style'' that is largely independent of prompting for age or education level.

\begin{table*}[ht]
\centering
\begin{tabular}{lcccc@{}}
\toprule
Target           & \textbf{ChatGPT} & \textbf{GPT3}  & \textbf{Bigscience-T0} &  \textbf{flan-T5-xxl } \\ \midrule

 \multicolumn{5}{@{}l}{\textbf{Age}}\\
    11-years-old &  0.00   &  0.00   & 2.00 &  3.00  \\
    11--12-years &  0.00    &  0.00    & 7.00 &  7.00   \\
   12--13-years  &  0.00    &  0.00    & 16.00 &  19.00   \\
   13--15-years & 0.00 & 4.00  & 16.00 & 13.00\\
   15--18-years &  0.00 &  7.00 & 19.00 & 13.00\\
   18--19-years &   0.00 & \cellcolor{gray!45}60.00 & 24.00 & 22.00\\
   22--23-years & \cellcolor{gray!45}100.00 & 36.00 & 6.00 & 10.00\\
  \midrule
     \multicolumn{5}{@{}l}{\textbf{Education}}\\
    5th graders &  0.00   &  0.00   & 2.00 & 5.00 \\
    6th graders &   0.00  &   0.00  &  11.00 & 12.00 \\
    7th graders &   0.00  &   0.00  &   11.00 & 15.00\\
    8th to 9th graders &  0.00   &  0.00   &  14.00  & 14.00\\
    10th to 12th grader &  0.00   &   4.00  &  11.00  & 9.00\\
    College students &   0.00  &  \cellcolor{gray!45}56.00   &  24.00 & 21.00\\
    College graduates &   \cellcolor{gray!45}94.00  & 43.00    &  6.00  & 2.00\\
    Professional &  0.00   &  4.00   & 2.00  & 2.00\\
    
  \midrule
   \multicolumn{5}{@{}l}{\textbf{Difficulty}}\\
   Easy explanation &   0.00  &   0.00  & 32.00 &  37.00\\
   Medium-level &   0.00  &  1.00   & 20.00 & 15.00 \\
    Difficult-level &  \cellcolor{gray!45}99.00   &   \cellcolor{gray!45}99.00  &  \cellcolor{gray!45}52.00 & \cellcolor{gray!45}55.00\\

    \midrule
   \multicolumn{5}{@{}l}{\textbf{Group}}\\
   kids &  0.00   &  0.00   & 6.00 & 3.00 \\
   Adults &  0.00   &  0.00  & 14.00 & 15.00 \\

       \midrule
   \textbf{Average} &  14.65   &  15.70   & 14.75 & 14.60 \\

  \midrule
                          
\end{tabular}
\caption{Percentage of responses in valid FKRE index range for the intended/prompted audience by model.\\ Performance above 50\% is highlighted in grey.}
\label{tab:param-models}
\end{table*}

Tables~\ref{tab:target-age},~\ref{tab:target-edu}, and~\ref{tab:target-group} show the detailed average FKRE scoress for each target group. In all three tables, the score should go down as we progress from left to right.
The \textit{None} target column shows the score without any specific group prompting. 

Table~\ref{tab:target-age} shows that as the age range increases, only GPT3 slightly decreases the average FKRE metrics. 
When considering education levels as the target group, as shown in Table~\ref{tab:target-edu}, both GPT3 and flan-t5-xxl exhibit the expected trend in scores for most education levels. 
Table~\ref{tab:target-group} shows the FKRE scores when differentiating between prompts for kids and adults. It is worth noting that, as also indicated in Table~\ref{tab:param-models}, all models only successfully generate answers within the acceptable reference range for a few target groups. However, the correlation between target groups and metric changes remains a crucial aspect to consider.

\subsection{Performance Evaluation}
Table~\ref{tab:param-models} summarizes the models' performance in generating valid responses for the various target groups. We compute the percentage of responses whose average FKRE value was within the recommended range for each target. The table is grouped into those defined groups for easier reading. 

ChatGPT and GPT3 perform poorly for age groups. ChatGPT's answers are valid for 22 to 23-year-olds or college graduates but not for younger groups. While they did not produce any responses that were in range for professionals, that omission is less severe: it just indicates that the answers chatGPT generated were at a more accessible level than recommended for that group.
GPT3 fairs slightly better, though it does not produce anything for children under 13 or 10th grade. 
Both models are good where they are good -- and bad everywhere else. This behavior is reminiscent of high-precision, low-recall classifiers.

BigScience-T0 produces some amount of responses that are within each target group's FKRE range. However, in none but one of the groups does it generate more than 50\% of the answer in range (the exception being difficult responses).
Flan-t5-xxl performs equally well or often slightly better than Bigscience-T0 -- but with the same caveat.
These models behave similarly to high-recall, low-precision classifiers.

Despite their differences, the average performance of all four models is relatively similar, on average producing around 15\% of answers within the recommended target readability range. No matter how we interpret that number and how it was computed, this level does not indicate that LLMs adapt well to the reading needs of different audience groups, even when explicitly prompted.

Or does it? To answer that question, we ran an additional test.


\subsection{Classifier Evaluation}
Our readability score analysis shows that the models, when prompted, do not consistently generate answers that adapt to the expected readability scores. 
There is, of course, the possibility that reading scores are ill-suited to the task, that the range does not apply to the kind of questions we have or other reasons. 
The metrics only focus on surface-level, count-based features and are designed for a different type of text than what we use. Consequently, they might misqualify a suitable LLM response by assigning a wrong score.
So despite the previous results, readability metrics may not capture all actual differences in the texts produced by LLMs. 

While these possibilities seem slim given the widespread use of these metrics in education, we validate this hypothesis (that the metrics do not capture all actual differences in texts).
To test this hypothesis and to strengthen our findings, we evaluate the results from the previous section in a different way, by classification. 

We assume that if there is a strong enough signal in the text for each of the target categories in our study, then a classifier should be able to predict the target age, education level, or group from the output text with some accuracy. This signal might be independent or immeasurable by the features used in FKRE.
In that case, we can accept that there are additional, measurable differences in the text that are not captured by Flesch-Kincaid's readability metrics. 
However, note that such a classifier could still not tell us exactly what those differences are.

We fine-tune a BERT model to classify the LLM responses into the target groups we used. We use BERT as it captures text meaning in a broader sense without us having to define specific features. We do not want to use an LLM, which we evaluate, but a simpler model.
We use 8,400 instances from each model: 6,720 for training, and 1680 for tests. We run over 5 epochs, with batch size 8, and learning rate=2e-5.

While classifiers can reliably distinguish between binary answers for kids and adults (F1 of 0.95), they fail to distinguish more fine-grained distinctions (i.e., age groups and education levels).\footnote{F1 was below 0.05, so we omit further details for space reasons} The detailed data for binary F1 and accuracy is in Appendix~\ref{appendix:f1}.



\section{Discussion}
Overall, our results show that current models fail to \emph{reliably} adapt their output to different audiences when prompted for age, education, or difficulty level. 
The findings reveal a striking lack of consistency and effectiveness in tailoring the generated responses to suit the intended target audience. In fact, the probability of a model generating an answer that is truly appropriate for the specified audience according to the reference range of metrics is a mere 0.15. This highlights a considerable gap in either the model's ability to accurately understand and cater to the unique needs and comprehension levels of diverse user groups or the lack of performance of the selected readability metrics. These results underscore the need for further advancements in model development and fine-tuning to ensure more reliable and effective adaptation to different audience segments. 


When evaluating the output of language models, especially in the context of LLMs, the reliability of the metrics must be considered. The metrics used to evaluate text quality and readability are typically designed for specific text lengths and types and are often based on assumptions about human-authored content. LLMs outputs are generated by machine learning approaches, and there may be a variation in the text length, structure, and coherence of the answer in comparison to the human-generated text. This divergence poses a challenge when applying conventional metrics designed for human-authored content and raises an open issue to explore more machine-generated text readability metrics.

\section{Related Work}
\label{sec:rel}

With the continuous advancements in systems for natural language understanding and generation, researchers in the field of education have been increasingly interested in exploring the potential of personalization and adaptive conversations to improve learning experiences~\cite{rooein2021adaptive, BAHA2022397, hong2023impact, kasneci2023chatgpt}. In addition, studies on text simplification help with modifying a sentence to enhance its readability and comprehension by reducing its lexical and syntactic complexity while preserving its core meaning.~\cite{martin2020muss} introduce a Multilingual Unsupervised Sentence Simplification system that eliminates the reliance on labeled simplification data. It utilizes sentence-level paraphrase data, unsupervised pretraining, and controllable generation mechanisms. 

LLMs, as ChatGPT, can serve as valuable allies for educators across all educational levels. They effectively enhance student engagement and accessibility by offering prompt and scalable answers to questions formulated in widely spoken languages. 
\citet{ROOEIN2023TAL} conducts a user study with teachers and students for using traditional chatbots in their classroom, and among the viability and potential effectiveness of the chatbot in education, it reports mixed sentiments, with some users expressing \textit{``I don't like chatbots!''}. It raises the question of whether introducing LLMs, particularly instruction-based models, can alter this scenario and deliver adaptive responses to users.

~\citet{cottonchatting} explores the potential advantages and challenges of implementing ChatGPT in higher education. In comparison, ~\citet{king2023conversation} specifically addresses the topic of plagiarism and its relationship with ChatGPT in the context of higher education. The study reports directly generated answers by the ChatGPt prompting by the author, including specific prompts about how a college student or college professor can use it for plagiarism.

\citet{murgia2023children} conduct a user study with 4th-grade pupils to measure the readability of ChatGPT-generated responses. Their findings suggest that ChatGPT can adapt responses, aiding children in comprehending curriculum-related concepts. However, the study also indicates areas where further improvement is needed.~\citet{haver2023use} evaluates the quality and readability of answers to cancer-related questions generated by ChatGPT, comparing them with answers provided by the National Cancer Institute (NCI).
The study reports FKGL and word count metrics on data for its readability metrics. Their qualitative assessment by experts shows the overall agreement for accuracy for NCI's answers is 100\% and 96.9\% for ChatGPT outputs for questions; however, it shows a few noticeable differences in the number of words or the FKGL score of the answers.

\section{Conclusion}
As LLMs are used increasingly in education, we need to know how well these models can adapt to different audiences. Specifically, we test how well four of the most commonly used models (two commercial and two open-source) adapt to a range of age groups and education levels. We evaluate the appropriateness of the responses via their average readability scores.
Our main finding is that models have a set ``target'' audience in terms of ease-of-readability, which is not hugely affected by prompting for specific age or education groups. Some models generate some answers within the readability range for each target, others generate all content in the same range. 
Our results suggest that current LLMs can not adapt well to the reading needs of specific groups, and are not particularly suited for educational purposes.

\section*{Ethical Considerations}
Age is a protected category. However, it is inextricably linked to the study of education in most cultures. In our study, we do not use data from actual people, but evaluate the age-appropriateness of a generated text via automated reading scores. There is such minimal risk of abuse and no concerns for the welfare of human subjects.

\section*{Limitations}
Our study uses automatic readability metrics. However, learning and education are about much more than ease of reading (or just reading in general). As such, our study can only outline some of the issues with current LLMs. Questions about other factors that affect learning, like appropriateness of style, individual tailoring, use of pedagogic concepts like encouragement, etc., are well beyond the scope of this work.

FKRE and FKGL are desigend for the US education system and English. They are unlikely to translate without adaptation to other countries, languages, and education systems.


\bibliography{custom}
\bibliographystyle{acl_natbib}

\clearpage
\appendix

\section{Appendix}
\label{sec:appendix}

\subsection{Questions}\label{appendix:questions}
What is the scientific method and why is it important?, What is the difference between a theory and a hypothesis?, What is the structure of an atom and how do atoms combine to form molecules?, What are the different types of cells in the human body and what are their functions?, What are Newton's laws in dynamics?, What are the fundamental laws of physics and how do they govern the behavior of the natural world?, How do scientists study the composition and behavior of different types of stars in the universe?, How does the process of natural selection work and how does it explain the evolution of species over time?, What are the different types of energy and how are they converted from one form to another?, What is the relationship between electricity and magnetism and how do they interact with one another?, How do scientists measure and track changes in the Earth's climate over time?, What is the greenhouse effect and how does it contribute to climate change?, How do scientists study the behavior of particles at the subatomic level using particle accelerators and detectors?, How do neurons in the brain communicate with one another and what role do neurotransmitters play in this process?, How do different types of lenses and mirrors work and how are they used in telescopes and microscopes?, What are the different types of biomes and what are the unique features and characteristics of each one?, How do scientists study the behavior and movements of animals in their natural habitats?, What is the relationship between genetics and behavior and how do scientists study this relationship?, What are the different types of rocks and how are they formed through geological processes?, How do scientists study the composition and behavior of the different layers of the Earth's atmosphere?, What is the process of photosynthesis and how does it enable plants to produce oxygen and food?, How do different types of viruses and bacteria cause disease in the body and how do scientists study and treat these illnesses?, What is the role of hormones in the body and how do they regulate various bodily functions?, How do scientists study the behavior and interactions of different species in an ecosystem?, What are the different types of biotechnology and how are they used in fields like medicine and agriculture?, How do scientists study the behavior and properties of light and what are the different types of electromagnetic radiation?, What is the relationship between plate tectonics and geological processes like earthquakes and volcanic eruptions?, How do different types of materials conduct and store electricity and how are they used in electronic devices?, What is the role of enzymes in digestion?, What is the difference between a hypothesis and a theory?, How do scientists classify living organisms?, How do scientists determine the age of rocks and fossils?, What are the fundamental principles of physics and how do they explain the behavior of matter and energy in the natural world?, What is the relationship between genetics and inheritance and how do scientists study these concepts?, How do vaccines work and how have they impacted public health?, What are the different types of energy and how do they affect the environment? What is the structure of the atom and how does it explain chemical reactions?, How do plants convert sunlight into energy through photosynthesis?, What are the different types of cells in the human body and what are their functions?, How does the human body maintain homeostasis and what happens when this balance is disrupted?, What are the different types of waves and how do they interact with matter?, What is the greenhouse effect and how does it contribute to climate change?, How do scientists study the deep ocean and what have they discovered about life in these extreme environments?, What is the history of the universe and how have scientists pieced together this timeline?, How do scientists study the behavior of subatomic particles and what have they learned about the nature of matter and energy?, How does DNA replication work and what are the different types of mutations that can occur?, What is the difference between renewable and non-renewable resources and how do we use them to meet our energy needs?, How do scientists study the human brain and what have they learned about its structure and function?, What are the different types of waves in the electromagnetic spectrum and how do we use them in everyday life?, How do ecosystems function and what are the different components that make up these complex systems?, What is the impact of human activity on the environment and what steps can we take to reduce our ecological footprint?, What are the different types of rocks and how do they form?, \\
What are the different types of chemical reactions and how do they influence the behavior of matter and energy?, What is the role of hormones in the human body and how do they regulate various physiological processes?, How do scientists study the effects of drugs on the human body and what are the different approaches to drug development?, How do geologists study earthquakes and volcanoes and what are the different types of geological hazards?, What is the role of symbiosis in the natural world and how do different organisms benefit from these relationships?, What is the impact of air pollution on human health and what steps can we take to reduce our exposure to harmful pollutants?, What are the different types of renewable energy and how are they used to generate electricity?, How do scientists study the behavior of matter at the atomic and subatomic levels and what are the different tools they use? What are the basic principles of physics?, How does the human brain work?, What is the process of photosynthesis?, What are the different types of cells in the human body?, How does DNA work?, What is the greenhouse effect?, What is the difference between a solid, liquid, and gas?, How do plants reproduce?, How do animals communicate with each other?, What is the difference between renewable and non-renewable energy sources?, What is the theory of relativity?, How do telescopes work?, What is the theory of evolution?, How does the circulatory system work?, What is the difference between a virus and a bacteria?, How do earthquakes happen?, What is the difference between weather and climate?, How does sound travel?, What is the process of mitosis?, How do airplanes fly?, How do we measure the age of the earth?, What is the process of fermentation?, What are the different types of clouds?, How do tectonic plates move?, What are the different types of fossils?, How do we measure temperature?, What is the process of meiosis?, What are the different types of rocks?, How does the human immune system work?, What is the difference between a physical and chemical property?, How do we use sound waves to diagnose medical conditions?, What is the process of osmosis?, What are the different types of biomes?, How do we use DNA technology to identify people?, What is the difference between a conductor and an insulator?, How do we use radioactive isotopes in medicine?, What is the process of evaporation?, How do we measure the strength of earthquakes?, What are the three states of matter and how do they differ from one another?

\subsection{T-test results}\label{appendix:ttest}
Results for t-test to examine whether there is a statistically significant difference in the means of the FKRE index between the None target group and the other target groups. 

\begin{table}[h!]
\centering
\begin{tabular}{lrr}
\toprule
Target                  & Statistic & P-value  \\ 
\midrule
11-years-old                 & -0.088031 & 0.929941 \\
11--12-years-old          & 0.660349  & 0.509797 \\
12--13-years-old          & -0.095860 & 0.923728 \\
13--15-years-old          & 0.351562  & 0.725540 \\
15--18-years-old          & -0.610243 & 0.542401 \\
18--19-years-old          & -0.132679 & 0.894582 \\
22--23-years-old          & -0.377245 & 0.706395 \\
5th graders                  & -0.602829 & 0.547312 \\
6th graders                  & -0.469268 & 0.639394 \\
7th graders                  & -0.689447 & 0.491349 \\
8th--9th graders   & 0.546740  & 0.585173 \\
10th--12th graders & -1.004327 & 0.316447 \\
college graduates            & 0.107010  & 0.914889 \\
college students             & -0.156879 & 0.875500 \\
professional                 & -0.325990 & 0.744776 \\
difficult-level explanation  & -0.105804 & 0.928155 \\
medium-level explanation     & -0.109604 & 0.912835 \\
easy-level explanation       & -0.744581 & 0.457408 \\
kids                         & 0.390140  & 0.696853 \\
adults                       & 0.397048  & 0.691759 \\
\bottomrule
\end{tabular}
\end{table}

\subsection{Classifier Evaluation}\label{appendix:f1}
The classification report with f1-score. 
\begin{table}[ht]
\centering
\begin{tabular}{lcc}
\toprule
 & F1-score \\
\midrule
ChatGPT & 0.953571\\
GPT3 & 0.95688\\
flan-T5-xxl & 0.953571 \\
Bigscience-T0 & 0.951786 \\
\midrule
\end{tabular}
\caption{Classification F1-score for the binary}
\label{tab:f1score}
\end{table}

\end{document}